MORE ROMANIAN WORD EMBEDDINGS FROM THE RETEROM PROJECT

VASILE PĂIȘ, DAN TUFIȘ

*Research Institute for Artificial Intelligence, Romanian Academy*
*{pais,tufis}@racai.ro*

## Abstract

Automatically learned vector representations of words, also known as "word embeddings", are becoming a basic building block for more and more natural language processing algorithms.

There are different ways and tools for constructing word embeddings. Most of the approaches rely on raw texts, the construction items being the word occurrences and/or letter n-grams. More elaborated research is using additional linguistic features extracted after text preprocessing. Morphology is clearly served by vector representations constructed from raw texts and letter n-grams. Syntax and semantics studies may profit more from the vector representations constructed with additional features such as lemma, part-of-speech, syntactic or semantic dependants associated with each word.

One of the key objectives of the ReTeRom project is the development of advanced technologies for Romanian natural language processing, including morphological, syntactic and semantic analysis of text. As such, we plan to develop an open-access large library of ready-to-use word embeddings sets, each set being characterized by different parameters: used features (wordforms, letter n-grams, lemmas, POSes etc.), vector lengths, window/context size and frequency thresholds.

To this end, the previously created sets of word embeddings (based on word occurrences) on the CoRoLa corpus (Păiș and Tufiș, 2018) are and will be further augmented with new representations learned from the same corpus by using specific features such as lemmas and parts of speech. Furthermore, in order to better understand and explore the vectors, graphical representations will be available by customized interfaces.

## *1. Introduction*

Word embeddings represent dense representations of words in a low dimensional vector space. Frequently there is made a contrast between the localist representations (the word themselves) and the distributed representations (the word embeddings). They are vectors of real numbers that can be used in natural language processing tasks instead of the words themselves, in order to improve the accuracy of different algorithms. This is happening because the vectors are capturing in some way the sense of the associated words. Thus, two words having similar meaning will have vectors with a small corresponding distance between them, as opposed to words with completely different meaning which will have distant vectors.

Modern approaches for computing vector representation of words make use of a famous observation made by linguist John Rupert Firth. According to this, "you shall know a word by the company it keeps" (Firth, 1957).

Distance between two vectors associated with words is usually computed using the "cosine distance". This is also known as the "cosine similarity", and uses the following formula:

$$CosineDistance(X,Y) = \frac{X \cdot Y}{||X||||Y||} = \frac{\sum_{i=1}^{n} X_i Y_i}{\sqrt{\sum_{i=1}^{n} X_i^2}\sqrt{\sum_{i=1}^{n} Y_i^2}}$$

where X,Y are the two vectors having dimension n, and Xi, Yi represent the i-th element of the corresponding vectors.

Current state of the art methods for obtaining word embedding representations are using artificial neural networks that are trained to predict one or more words, given for input a sliding window of context words from the training corpus. This approach has the advantage of being completely unsupervised, requiring only a large training corpus in order to produce the representations. This approach was initially introduced by Bengio et al. as described in (Bengio et al., 2003).

If the neural network is being trained to predict a middle word given a context window of m words, the model is referred to as CBOW ("Continuous Bag Of Words"). On the other hand, if the neural network is trying to predict the context based on a given central word, the model is referred to as Skip-gram. This is detailed in the paper by Mikolov et al. (Mikolov et al., 2013). Both methods can produce comparable representations and these are actually the weights learned by the network. Vector length does not depend on the vocabulary size, being a network hyper-parameter (equal to the number of neurons on the hidden layer of the network, orders of magnitude smaller then the vocabulary). Contrary to other co-occurrence-based representations that depend on vocabulary size, word embeddings are dense representations in much less dimensional spaces, thus less computationally demanding.

In the ReTeRom project we are aiming to develop an integrated and configurable chain of advanced language processing modules, based on automated learning, for morphological, syntactic and semantic analysis of Romanian texts. The underlying data-source is the CoRoLa corpus, on which various algorithms were trained and tested. The new resources that will be created in the ReTeRom project will be added to the CoRoLa bimodal corpus.

The Reference Corpus for Contemporary Romanian Language (CoRoLa) (Barbu Mititelu, 2018) was constructed as a priority project of the Romanian Academy, between 2014 and 2017. It contains both written texts and oral recordings. Its aim was to cover major functional language styles (legal, scientific, journalistic, imaginative, memoirs, administrative), in four domains (arts and culture, nature, society, science) and in 71 sub-domains while taking into account intellectual property rights (IPR). With over 1 billion word tokens (written and spoken), CoRoLa (corola.racai.ro) is one of the largest fully IPR-cleared Reference Corpus in the world. CoRoLa is searchable via three interfaces (two for written part and one for the oral part) and it is supported by the KorAP corpus management platform, developed at Institute for German Language in Mannheim, (Bański et al., 2014), (Diewald et al., 2016).

Apart from the raw texts and sound records, the CoRoLa corpus contains linguistic annotations in the form of: phoneme, syllable, lemma, text-to-sound alignments, part of speech (POS) tagging, syntactic chunking and dependency parsing. Several of these annotation types were added using the TTL (Tufiş et al., 2008) service.

In a previous work (Păiș and Tufiş, 2018) we focused on computing word embeddings on the CoRoLa corpus, using the complete word forms. In the mentioned paper we provided various quantitative and qualitative details and compared the results with previous state of the art known vectors (Bojanowski et al., 2016), (Pre-trained word vectors using Wikipedia corpus), obtained by Bojanowski and his colleagues using the Romanian Wikipedia corpus. We demonstrated that the CoRoLa based representations are superior to other known ones for the Romanian language. Therefore, it was natural to select these representations for use in the ReTeRom project.

## *2. Visualizing Romanian word embeddings*

One problem we were faced with when thinking of ways to evaluate the quality of the learned representations was the difficulty to interact with the model in its vector format. On the other hand, integrating the vector representations into algorithms requires adequate means for retrieving, selecting, filtering and visualization of the word embeddings of interest. In the following we will describe the existing interfaces by means of which a user can investigate the significance of the numerical encodings of words and their used contexts.

Our previous evaluations indicated that the best representation using word forms from the CoRoLa corpus was based on vectors with 300 dimensions. The interrogation interface constructed previously allowed for asking the model to provide *n* similar words to a certain given word, or to ask a question in the format "what word is similar to word C in the same way as word B is similar to word A?". This is also known as analogies. Using vector notation, this question can be formulated as "A-B+C". Such a question with its associated answer is:

*vec("rege")-vec("bărbat")+vec("femeie")=vec("regină")*

This can be formulated as: what word is similar to "femeie" ("woman") in the same way as "bărbat" ("man") is similar to "rege" ("king"). In this case, the answer is the expected one "regină" ("queen").

Figures 1 and 2 shows examples of the two textual query interfaces.

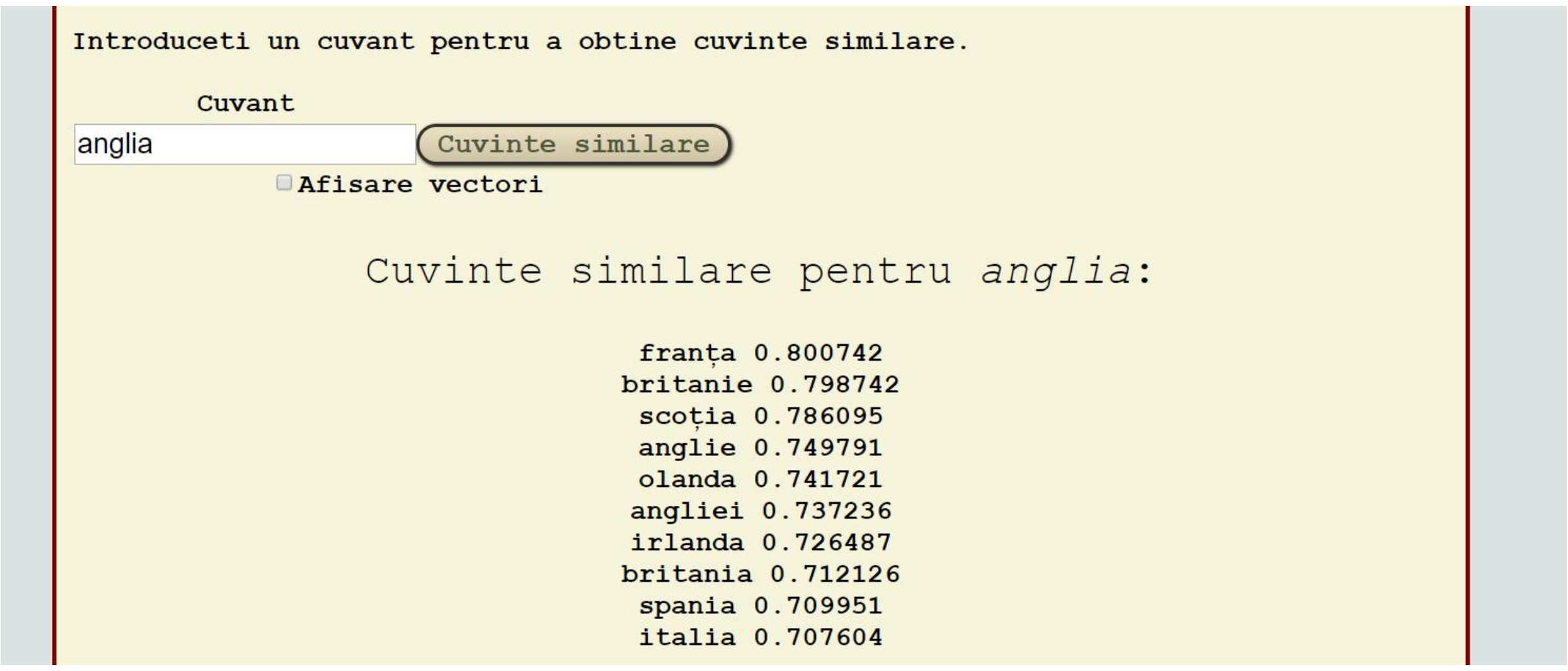

**Figure 1.** Textual interface for querying similar words

In Figure 1, the (10) words most similar to "anglia" are shown with the similarity score (cosine distance). Figure 2 shows a snapshot illustrating the linear algebraic properties of the vectors, corresponding to the query: *vec("anglia")-vec("londra")+vec("paris") = ?.* The result, *vec("franța")*, signifies that the vector for the word "franța" is the closest (in terms of cosine distance) to the vector resulting from the simple algebraic operations *A-B+C.* At the bottom of the Figure 2 there are displayed the vectors for the words in the query, the partial results of the query (A-B, A-B+C), the vector of the word answering the questions (A-B+C-R) as well as the distance motivating the answer (cos (A-B+C;R)).

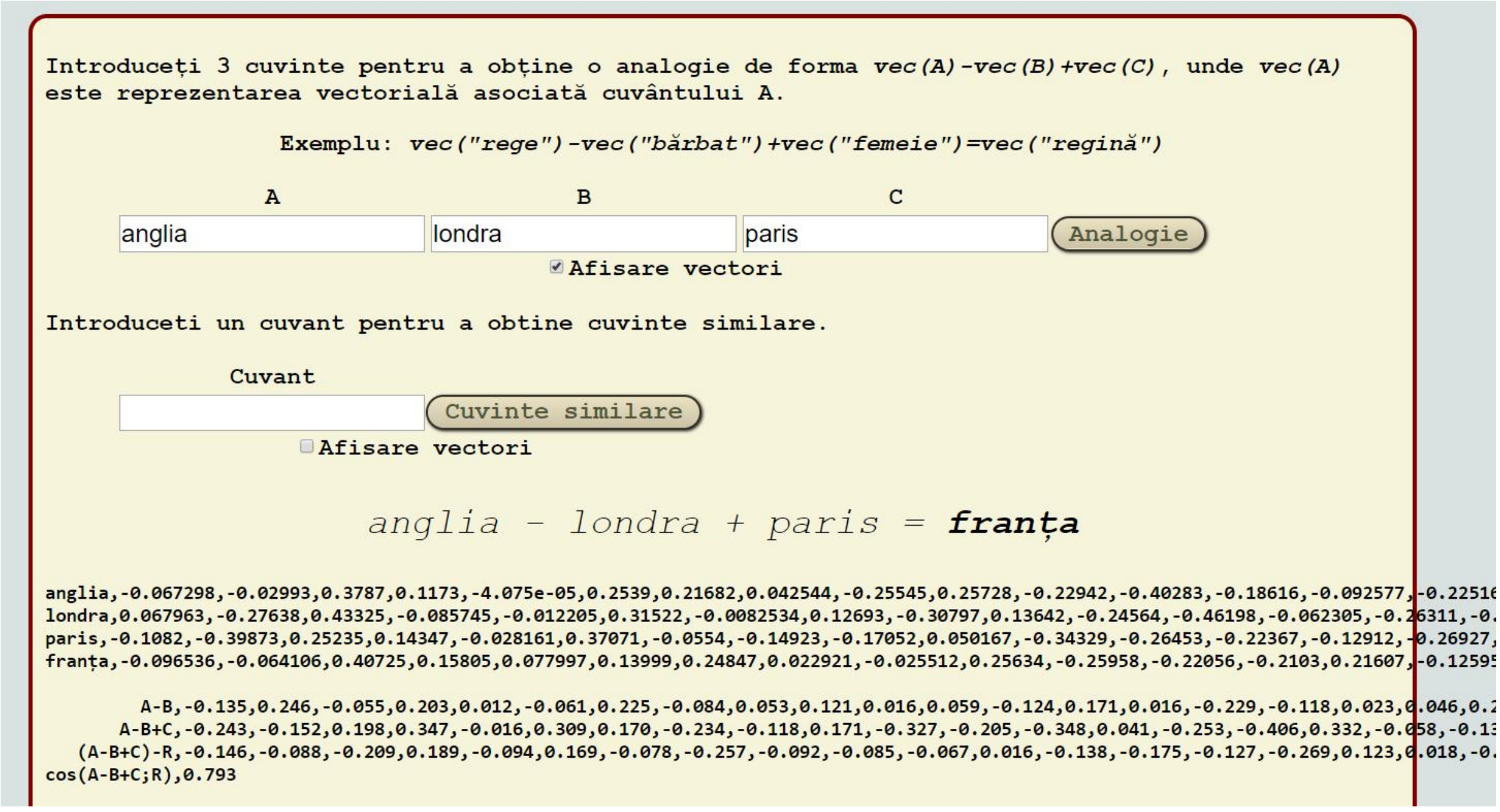

**Figure 2**. Textual interface for querying analogies

Nevertheless, using such simple textual interfaces makes it hard to understand results comprised of more words, like the most similar 1000 words. For this reason, we decided to construct additional visualization interfaces for the use of ReTeRom project, using graphical approaches. The difficulty of constructing graphical interfaces for visualizing word vectors is given by their size. Thus, dimension reduction techniques must be used to project the vectors from dimension 300, as those used in the ReTeRom project, to the two dimensions that can be rendered on the computer screen.

The first interface constructed used the t-Distributed Stochastic Neighbor Embedding (also known as t-SNE) technique for dimensionality reduction, as described in (van der Maaten and Hinton, 2008), (van der Maaten, 2014). Since one of the objectives of the ReTeRom project is to obtain a web-based platform for interacting with the various tools and resources at project's disposal, we decided to use a JavaScript based implementation of the t-SNE algorithm, as described in (tSNE-JS). This was adapted to use the learned representations based on the CoRoLa corpus.

One implementation allows representing for the most frequent *n* words present in the corpus, with *n* being a parameter selectable by the user. An example for the most frequent 300 words is given in Fig. 3.

A second implementation focuses on representing *n* words most similar to a given word. In this way, the user has better control on what is displayed. An example is shown in Fig 4.

In Fig. 4 we used the same query as with the textual interface. However, this time it's easier to understand what kind of data is returned by the model. In this picture, in the lower right corner are grouped various countries (these were the results returned first in the textual query interface in Fig. 1). In the upper left corner of the picture, there are different cities and regions in England. In other parts of the image, it can be seen other related words, such as names of historical people ("Churchill", "Edward", "Stuart"). The interface based on the t-SNE algorithm is interactive, allowing the user to pan and zoom using the mouse, in order to better explore the generated image.

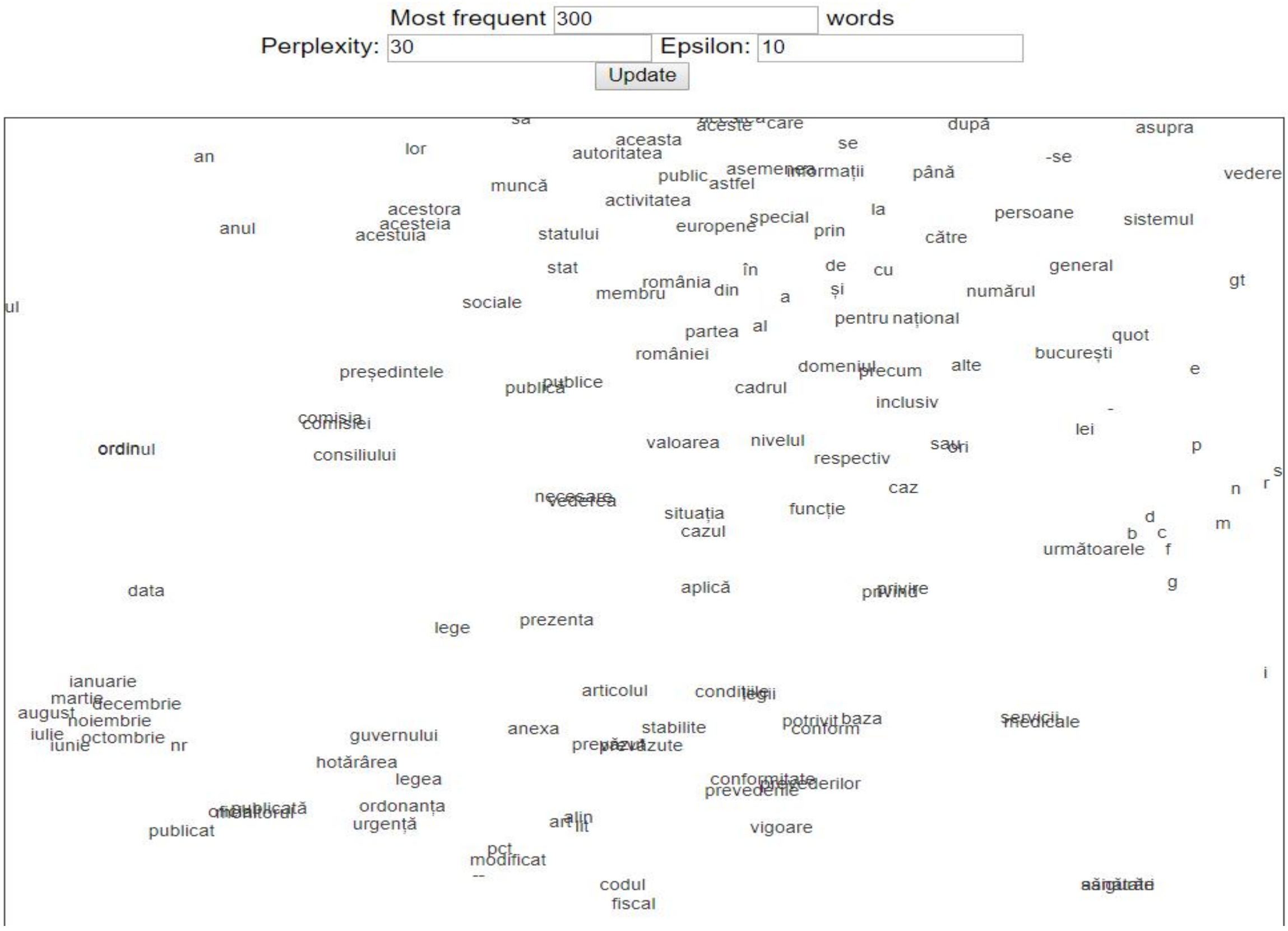

**Figure 3.** Using t-SNE to obtain a representation for the most frequent 300 words in CoRoLa.

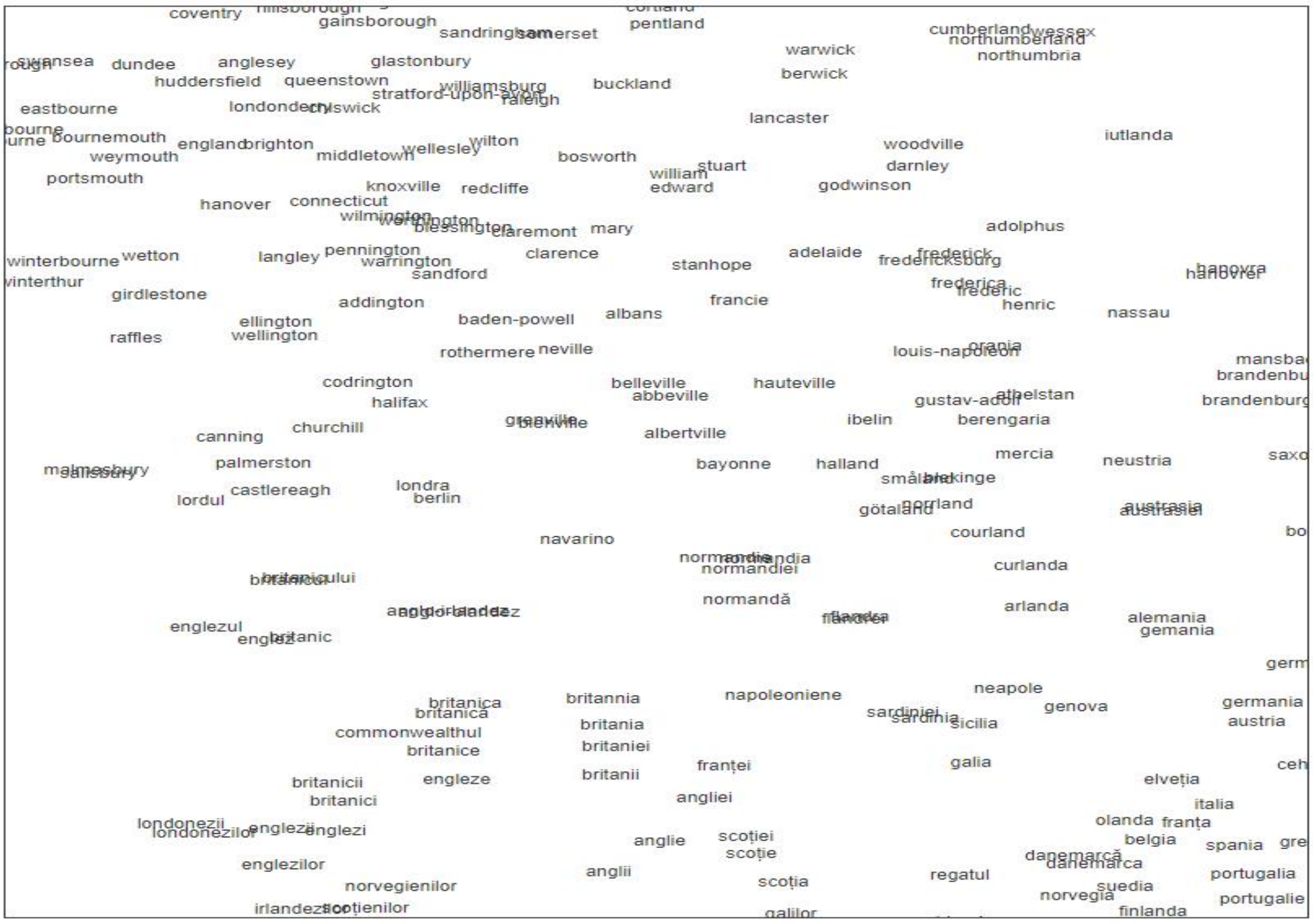

**Figure 4**. Using t-SNE based interface to show the most similar 300 words to the word „Anglia"

Another interface was constructed using a graph-based visualization method. In this case, the user starts by entering a word which will become the center of a tree and *n* most similar words will be brought by the system and connected to the given word. Afterwards the user can bring other similar words either by entering a new word or clicking on one of the words displayed. This will cause the system to load new nodes and connect them to existing ones, thus constructing a graph with the words as nodes and the similarity between them indicated by edges.

An example representation using this interface, for the word "motocicletă" ("motorcycle"), with 50 similar words, is given in Fig. 5.

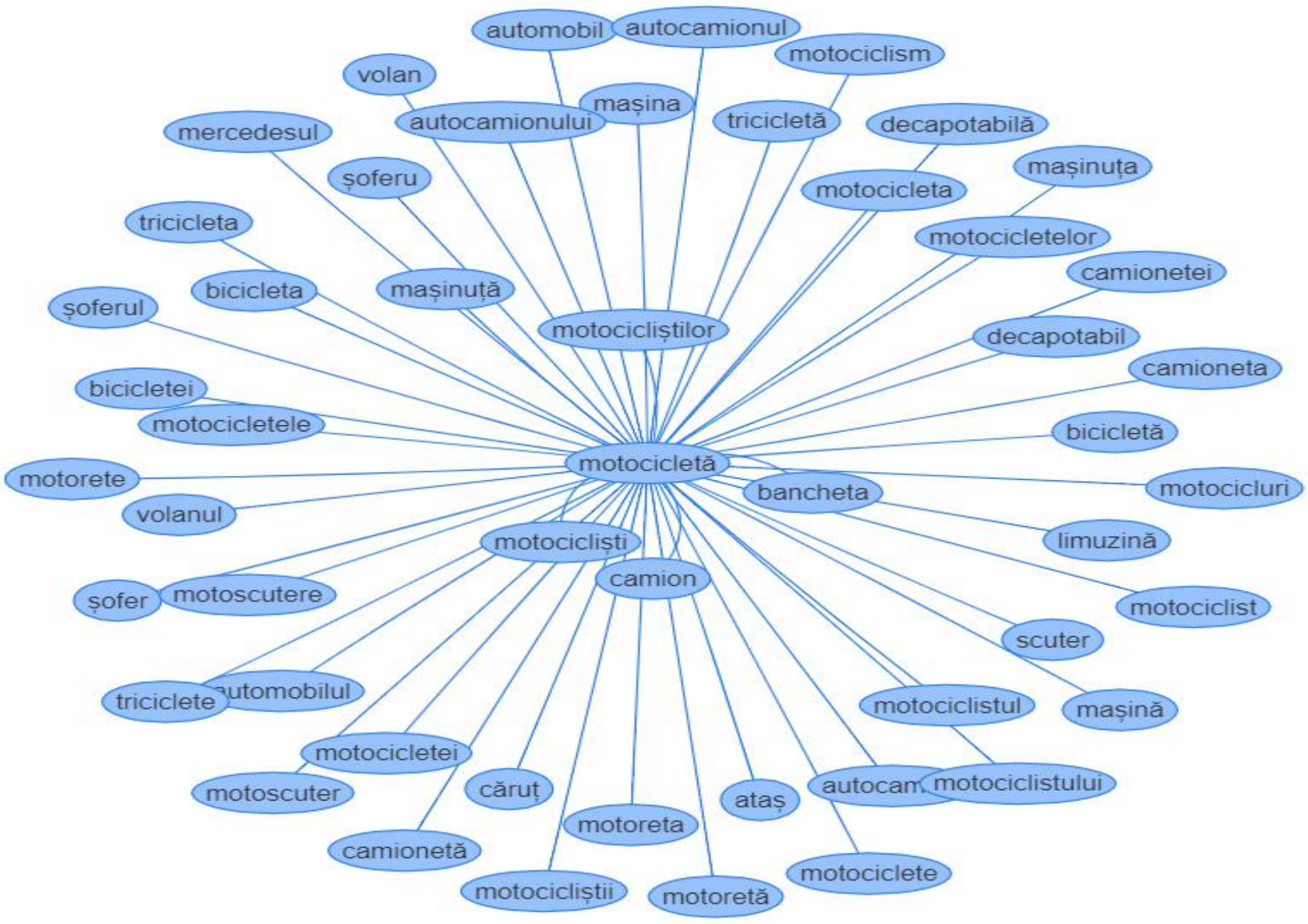

**Figure 5.** Graph based representation for the word "motocicletă" ("motorcycle") with 50 similar words

Since the user can manually add multiple words to the visualization it is envisaged its suitability for identifying similarity paths between multiple words. For example, you may wonder if there is a connection between "motorcycle" and "boxing". Apparently, there isn't any direct connection. However, by adding on the same diagram the two words "motocicletă" and "box" a connection becomes visible. Both words are similar in a way to the word "motociclism" which is a sport (similar to "box") which uses "motocicletă". This can be seen in Fig 6.

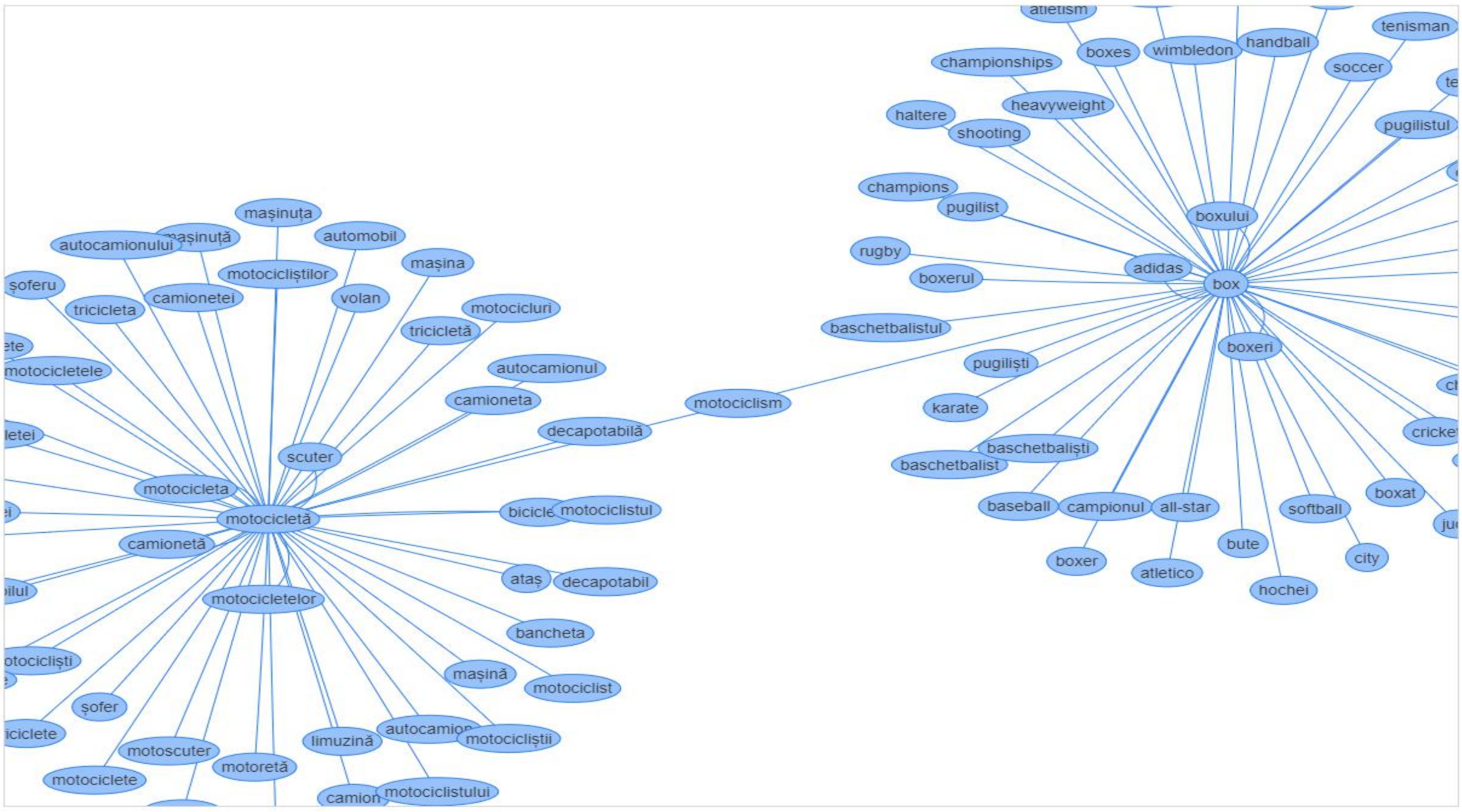

**Figure 6.** There is an indirect similarity connection between “motocicletă” and “box” illustrated in the graph-based interface

Graphical interfaces do not make obsolete the initial textual interfaces. Instead, they complement them since it’s possible to use the textual analogies interface to ask an analogy question and then the graphical interface can be used to better understand why a certain word was chosen as answer. The four words from Fig 2. can be added on the same diagram, thus obtaining additional related words. This can be seen in Fig 7.

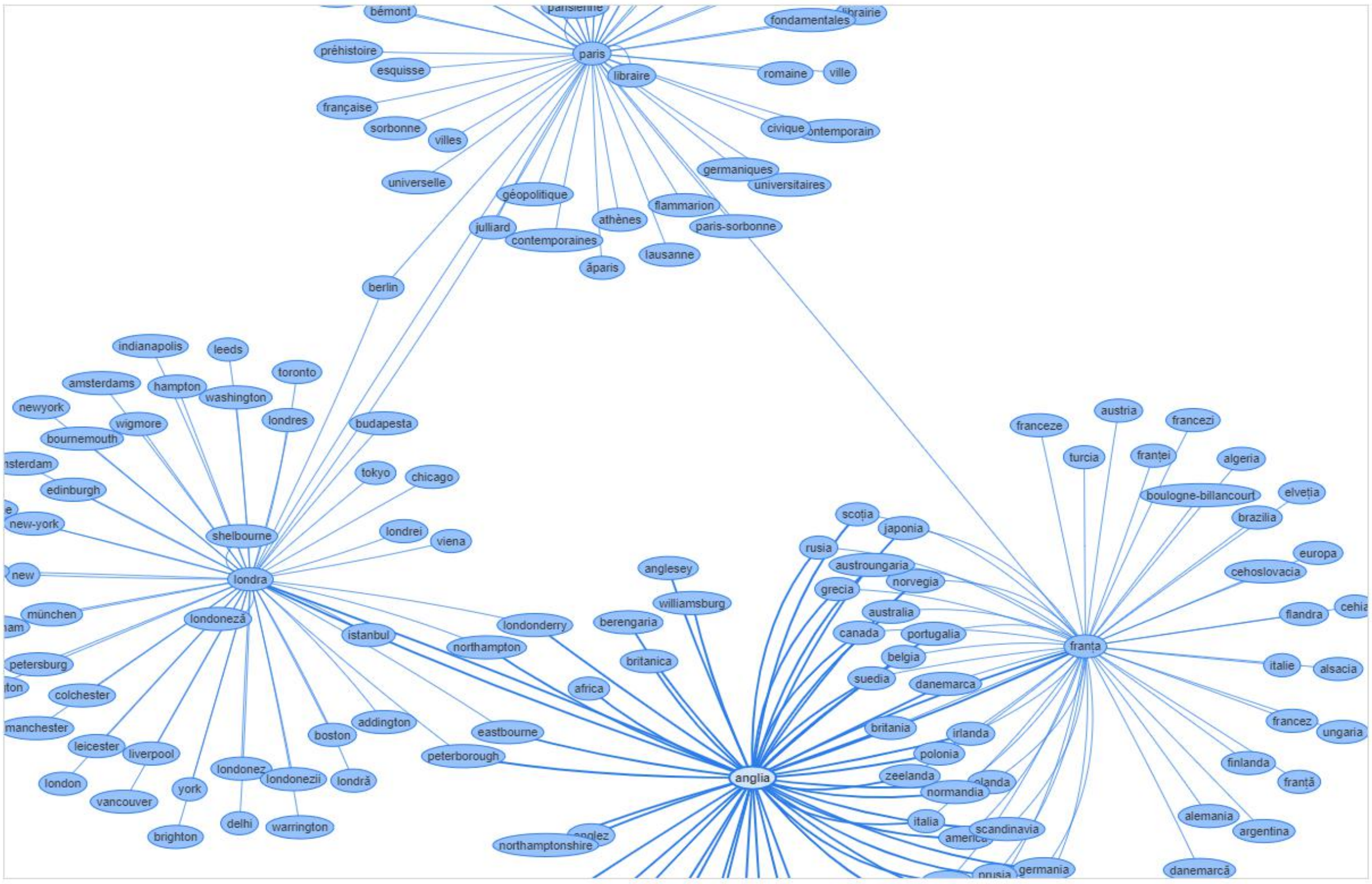

**Figure 7.** The words used in the textual interface in Fig. 2 shown on the graph-based interface

## *3. Necessity of other embeddings*

Since the embeddings reported in (Păiș and Tufiș, 2018) were produced using the actual word forms available in the CoRoLa corpus, various inflected forms of the words are part of the representation. Analyzing the visualizations, it becomes obvious that amongst the top similar words reported for a given word there are actually several of its own inflected forms.

The presence of inflected forms is not an issue when using the model for computing distances between known words. However, one of the envisaged uses in the ReTeRom project is to obtain similar words and not just computing distances between given words. Therefore, the presence of inflected forms must be addressed.

Since the CoRoLa corpus already contain lemmas as part of its annotations, it becomes an obvious choice to use them for constructing new embedding models.

We decided to keep the model dimensions the same (300) and build two models, one based on the actual case sensitive lemma, as produced by the TTL web service (Tufiş et al., 2008) and present in the corpus annotation, and another based on the lowercased lemma.

In the case of regular word embeddings, constructed based on the actual word form, as we did in (Păiș and Tufiș, 2018), different senses associated with the word are captured in the same vector. This happens because the vector is associated with the word form and not directly with a particular sense. It has been suggested that multiple senses of a word actually reside in linear superposition within the standard word embeddings (Arora et al, 2016). This idea is challenging and will be explored by future research.

For a model built on lemmas, there isn't only a superposition of senses but also a superposition of wordforms (since multiple words are associated with the same lemma).

The reason for having the two models was to try to see if a differentiation between lemmas corresponding to named entities, such as persons or organizations, and lemmas corresponding to other nouns is possible.

The previously constructed visualization interfaces (both the basic textual ones and the graphical ones) were adapted to be used with the new models. The statistics and evaluations for the new vectors will be largely described and commented in a forthcoming paper but here we only note that the results are quite different concerning the similarity vicinities in the lemma-based vector spaces. This is not surprising if one considers the above-mentioned superposition of the vectors attached to the words from a lemma inflectional paradigm. Yet the form of the superposition should be new, since different combinations of morphological features impose context restrictions as well as the different frequencies of various inflectional forms of a lemma contribute to the coordinates in the vector space of the vector associated to a given lemma. This is another topic of further research.

In Fig. 8 is presented a comparison of graph-based representations obtained from the word form model and the lemma-based model. As expected, inflected forms do appear in the first image ("manualul", "manual", "manualele"), but they do not appear in the lemma-based representation, thus allowing the user to easily identify other related words. Also, the similarity vicinities are quite different, with only the item ”manualitate” occurring in both vicinities shown in Figure 8.

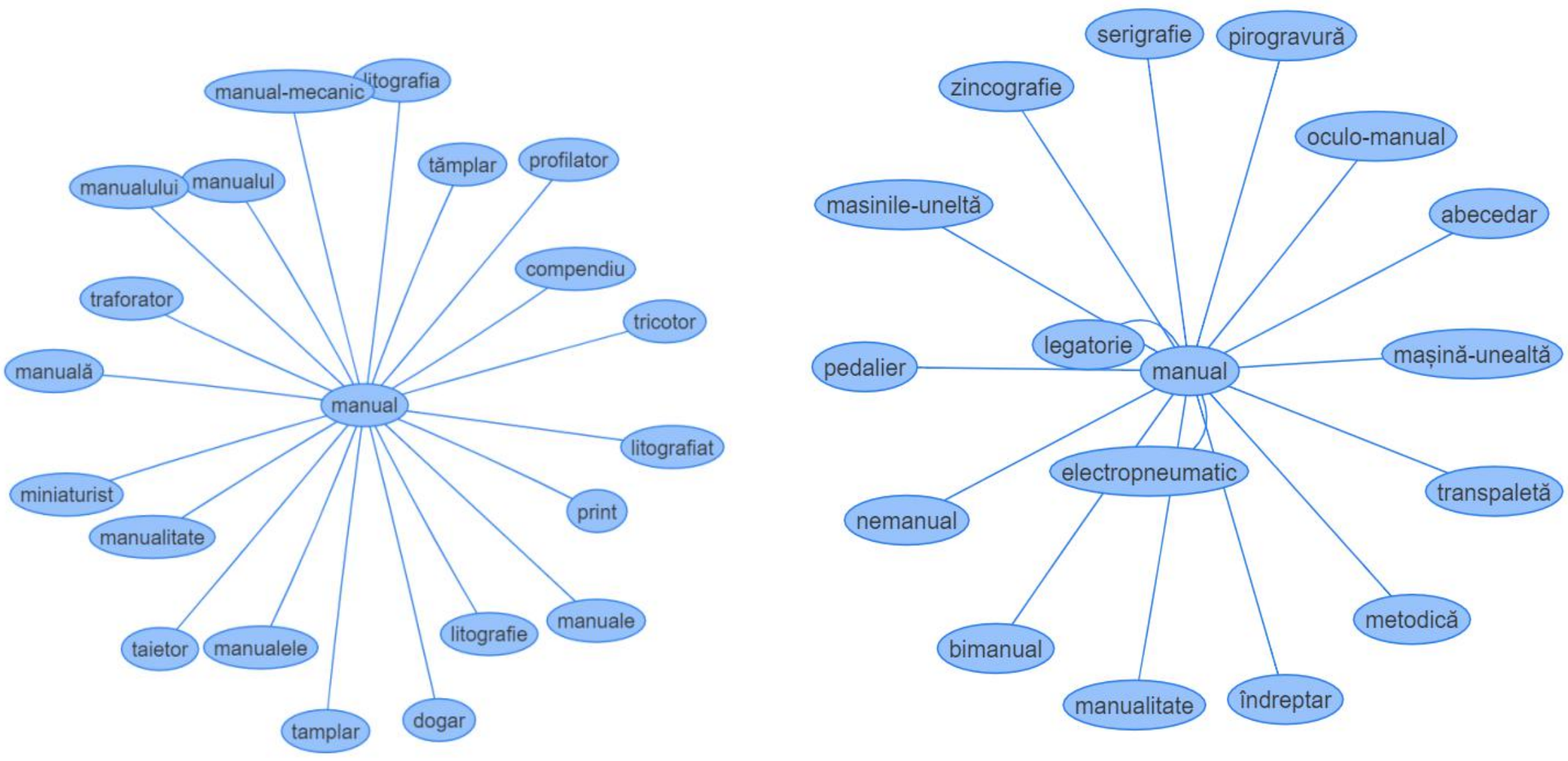

**Figure 8.** Comparison of representations obtained from the word form model (left) and the lemma-based model (right)

## *4. Accessing the vectors*

As mentioned before, we plan to develop a module (interface to the library of sets of word vectors) so that the appropriate vectors may be used in specific applications. The implementation of the module will follow the development conventions (Ion, 2018a), (Ion, 2018b) adopted for the ReTeRom text processing flow (full-rest) and it will offer programmable means to extract the needed vectors, characterized by used specified features: word, lemma and/or POS-based, vector length, frequency-threshold used for the vector construction, list of nearest similarity neighbors of a vector for a given word, etc.).

## *5. Conclusions*

Word embeddings are becoming a common way of representing the words in various natural language processing tasks. Therefore, for the purposes of the ReTeRom project, this kind of representation will become one of the resources used. Previously computed word embeddings on the CoRoLa corpus will be used, together with new representations learned using lemmas.

Furthermore, in order to visually explore these resources, several web based interfaces were constructed, using simple textual searches, t-SNE dimensionality reduction technique and a graph based approach for connecting visually similar words.

### *Acknowledgements*

This work is supported by a grant (73PCCDI/2018) of the Ministry of Research and Innovation CCCDI-UEFISCDI, project code PN-III-P1-1.2-PCCDI-2017-0818 (ReTeRom) within PNCDI III.